4
1
# Image similarity using Deep CNN and Curriculum Learning

Srikar Appalaraju  
srikara@amazon.com

Vineet Chaoji  
vchaoji@amazon.com

Amazon Development Centre (India) Pvt. Ltd.
abstract
**Image similarity involves fetching similar looking images given a reference image. Our solution called *SimNet*, is a deep Siamese network which is trained on pairs of positive and negative images using a novel online pair mining strategy inspired by Curriculum learning. We also created a multi-scale CNN, where the final image embedding is a joint representation of top as well as lower layer embedding's. We go on to show that this multi-scale Siamese network is better at capturing fine grained image similarities than traditional CNN's.**

*Keywords* — **Multi-scale CNN, Siamese network, Curriculum learning, Transfer learning.**
## I. INTRODUCTION

The ability to find a similar set of images for a given image has multiple uses-cases from visual search to duplicate product detection to domain specific image clustering. Our approach called *SimNet*, tries to identify similar images for a new image using multi-scale Siamese network. Fig. 1 shows examples of image samples from CIFAR10 [39] on which *SimNet* is trained on.

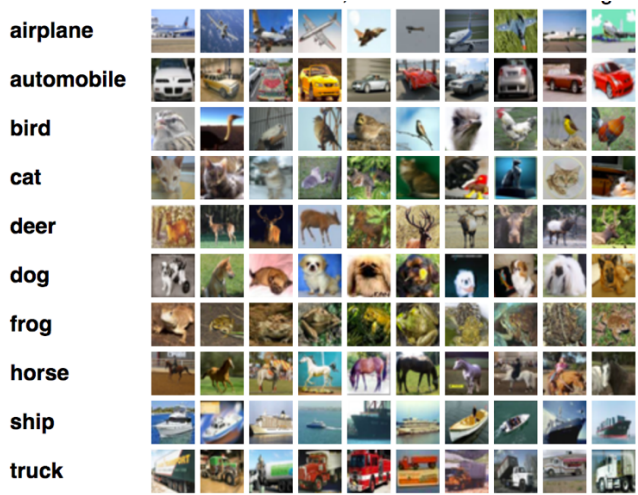

*Fig. 1 examples of CIFAR 10 images. Task is - given a new image but belonging to one of the 10 categories, find similar set of images.*

**Algorithms**: Image verification algorithms aim to determine whether a given pair of images are similar or not. Image verification is different from image identification. Former solves similar images use-cases whereas latter is more of an image retrieval nature. The advancements in Image verification field is in two broad areas – a) image embedding and b) metric learning based. In image embedding, a robust and discriminative descriptor is learnt to represent each image as a compact feature vector/embedding. Typical image descriptors include *SIFT* [1], *LBP* [2], *CNN* embedding's etc. Current [1]state of the art feature descriptors is generated by *CNN* which learns features on its own. In metric based learning, a distance metric is learned from labeled training samples in an embedding space to effectively measure the similarity of images.

*SimNet* uses a multi-scale CNN in a Siamese network which learns a 4096-dimensional embedding of an image. It learns a set of hierarchical nonlinear transformations to project images pairs into a 4096D subspace, under which the network tries to minimize distances between positive image matches and maximize for negative matches. Siamese network require pair forming i.e. positive image pairs (near similar images) and negative image pairs (non-similar images) for it to learn distance margin. Choosing the right pairs of images for training turns out to be very important for achieving good model performance and faster model convergence. We propose a novel online pair mining strategy (*OPMS*) which tries to ensure consistently increasing difficulty of image pairs as the network trains. This is inspired from Curriculum Learning [4]. With this we present the main contributions of this paper –

1. Multi-scale CNN used in a Siamese network. This CNN learns a joint image embedding of top as well as lower layers. This model learns a much better image embedding's than a traditional CNN for the task of image similarity.
2. We employ a novel online pair mining strategy inspired from Curriculum learning which ensures the model finds better local minima and faster model

---

[1] Accepted to GHCI 17 Oral talk in Artificial Intelligence, Data mining and Machine learning track

convergence with negligible drop in performance.

Overview of the rest of the paper is as follows: in section 2 we briefly discuss the data and the metrics used to evaluate models. In section 3 we present a brief intro to Siamese network, our custom multi-scale CNN architecture and online pair mining strategy. In section 4 we share model training and prediction process. In section 5, we explore related work. In section 6 we show the results and final hyper-parameters of our model. Finally, in section 7 we have summary.

## II. DATA USED

Training a supervised ML model needs images with labels. We use CIFAR10 [39], which is an established computer-vision dataset used for object recognition. It is a subset of the 80 million tiny images dataset [40] and consists of 60,000 32x32 color images containing one of 10 object classes, with 6000 images per class. The dataset is divided into five training batches and one test batch, each with 10000 images. The test batch contains exactly 1000 randomly-selected images from each class. The training batches contain the remaining images in random order, but some training batches may contain more images from one class than another. Between them, the training batches contain exactly 5000 images from each class.

All trained models are evaluated on accuracy. All model training and hyper-parameter tuning is done using 5-fold cross validation whereas test set is used only once at the end to report the final model performance. As it's a publicly available dataset, the dataset ensures the class distribution in train and test data were similar.

## III. SIAMESE NETWORK

We modelled the problem as one of image verification/similarity. For this we used a Siamese network architecture where we have 2 *CNN's* whose weights are shared and they are trying to minimize a loss function. Formally, a Siamese network is a function *f* that maps each image *I* into an embedding position *x*, given parameters $\theta$. $x = f(I; \theta)$. The parameter vector $\theta$ contains all the weights and biases for the convolutional and inner product layers, and typically contains 1M to 150M parameters depending on the size of the network. The goal is to solve for the parameter vector $\theta$ such that the embedding produced through *f* has desirable properties and place similar images nearby. See Fig. 2

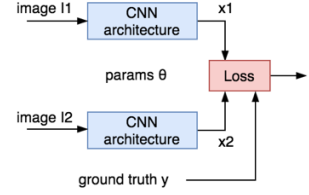

*Fig. 2 Siamese network architecture.*

The input to the network is pairs of images (see fig. 3) i.e. consider a pair of positive images ($I_q$, $I_p$) that are two views of the same image; got by data augmentation or two different variations of same category image, and a pair of negative images ($I_q$, $I_n$) that are from different categories. We can map these images $I_q$, $I_p$, $I_n$ through our network to get embedding's $x_q$, $x_p$, $x_n$. If the network had learnt a good embedding, we would find that ($x_q$, $x_p$) are nearby while ($x_q$, $x_n$) would be further apart.

The image embedding is got by a deep convolutional neural network. The network has multiple layers (*M*) and $n_m$ neurons in the $m^{th}$ layer, where m=1, 2, …, *M*. For a given image, $x \in \mathbb{R}^d$ the output of the $m^{th}$ layer is $h^m = s(W^m.x + b^m) \in \mathbb{R}^{pm}$ where $W^m$ is a projection matrix to be learnt in the $m^{th}$ layer and $b^m$ bias vector. s: $\mathbb{R}^d \to \mathbb{R}^{pm}$ is a non-linear activation function here ReLu [7]. Finally, we get a function *f*: $\mathbb{R}^d \to \mathbb{R}^{pm}$, a parametric non-linear function that projects an image of *d* dimensions into a sub-space of *p* dimensions in the $m^{th}$ layer. In this sub-space, we would like similar images to be closer to each other and dissimilar images to be further apart. See fig. 3

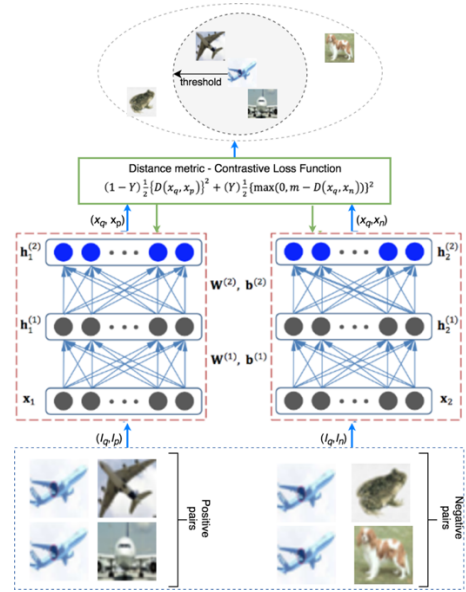

*Fig. 3 Siamese network. positive and negative pairs are the input to train the network. Once trained, similar category images are closer than dissimilar images in the 4096 D embedding sub-space.*

The network loss can be formalized as the contrastive loss function *L* [11] which measures how close *f* is able to place similar images nearby and keep dissimilar images further apart. For one training image, the loss is defined as shown in Eq.1. *m* kept at 1. Setting *m* to some value should not impact learning, distance metric *D* would simply scale accordingly.

Label of *Y=1* is given to dissimilar images or negative pairs and label of *Y=0* to similar images or positive pairs. The two *CNN's* of the Siamese network have shared weights optimized by contrastive loss function *L*.

$$L(\theta) = (1-Y)\frac{1}{2}\{D(x_q,x_p)\}^2 + (Y)\frac{1}{2}\{\max(0, m - D(x_q,x_n))\}^2 \quad (1)$$

*Eq. 1 Contrastive loss function (L). computes loss per training example. Total loss is summation over all image pairs. m=1 (value of m does not impact learning as D would simply scale accordingly).*

**Contrastive loss Function $L(\theta)$**: In Eq.1, when images are similar *Y=0*, the right-hand additive part goes away; the loss becomes the distance between two similar image embedding's i.e. if the training images are similar, we would want to reduce the distance between them which the network learns. When images are dissimilar *Y=1*, the left-hand additive term goes away; the right-hand term is basically hinge loss. The loss function value becomes 0 if the images are totally dissimilar (like cat and aircraft image hence no minimization is required. But if the images are somewhat similar (cat and dog images) then we do some minimization as there is an error. *m* is margin between positive and negative images. Value of *m* is empirically decided. Larger *m* pushes similar and dissimilar images further apart, *m* acts as margin. Here we have used *m=1*.

### A. Online Pair Mining strategy (OPMS)

In order to train a Siamese network we need pairs of images with label as input. In our case, we need to generate two types of pairs - positive pair ($I_q$, $I_p$) with label 1 and negative pair ($I_q$, $I_n$) with label 0 (see fig. 3). Consider image pairs ($I_q, I_p$) ($I_q, I_n$) $\in \tau$ where $\tau$ is the set of all possible pairs that can be generated from training images *I* having *M* categories. For these *M* categories, each with *n* variations on average (different image variations) the number of positive pairs that can be generated are $\sum_{i=1}^{M} \frac{n_i(n_i-1)}{2}$ and number of negative pairs are $\sum_{i=1}^{M}(I-n)n$. In our case, with *I* as 50K images and M as 10 categories and *n=1000* per category we get about 500 million image pairs; which results in extremely slow convergence. Plus, not all image pairs contribute to model learning equally. The main idea here is to select "hard image pairs" and introduce them gradually in training. This helps better model learning. On CIFAR 10, we found *SimNet* when trained via curriculum learning performed 18.8% better than when trained with random pairs (see table 2, rows B and F), which clearly shows the benefit of Curriculum learning and OPMS. Hard image pairs are near decision boundary as shown in fig. 4. In order for this to happen we introduce a) Pair constraint and b) use ideas from Curriculum learning.

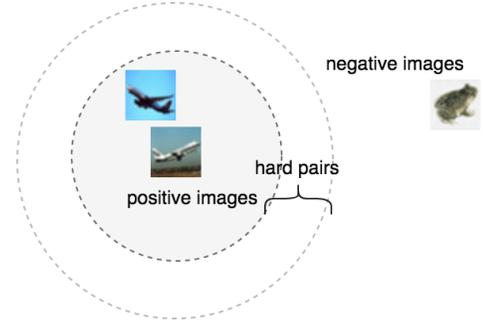

*Fig. 4 rough visualization for a particular (airplane) anchor image, the positive images decision boundary (first circle) and the images after the second circle are the negative images. The gap close to the boundaries is where the hard image pairs are.*

**Pair constraint:** Pair constraint (eq. 2) was devised such that only pairs which satisfy this constraint are considered for training. This means that given image $I_q$, we want to select image $I_p$ (hard positive) such that the distance between the embedding's of $I_q$ and $I_p$ is maximum; but at the same time select negative image $I_n$ (hard negative) such that the distance between embedding's of $I_q$ and $I_n$ is minimum while maintaining the inequality. *m* is margin between positive and negative pairs. Generally speaking, the hardest pairs would be the ones which fall at the edge of the decision boundary. Pair constraint helps remove false-positives from training which fall within the margin *m* or below.

$$argmax\{D(x_q,x_p)\}^2 + m < argmin\{D(x_q,x_n)\}^2 \quad (2)$$

The number of such pairs to be picked for training in each mini-batch is controlled by λ hyper-parameter with value between 0 and 1. Higher the value of lambda, tougher pairs would be picked for training. Idea is to create image pairs such that they generate some loss so that the network learns with each training pair. If for some pair the loss is zero, it's a waste of time as the network is not learning. Eq. 2 requires the computation of *argmin* and *argmax* for the whole training set which is infeasible. Additionally, even after doing so, might lead to poor training as mislabeled and poor images would dominate the hard

positive and negative pairs. Two possible ways to circumvent this issue –

1. Generate image pairs *online* as part of training itself in mini-batches.
2. Generate image pairs *offline* every *n* training steps on a subset of the training data.

In our implementation, we adopted the online approach and use large mini-batches to have good enough samples to satisfy Eq. (2). Offline approach required model checkpointing and dumping the pairs onto disk which is inefficient. In online approach, the generation of image pairs is done in *CPU* in parallel on multiple threads with *GPU* used for model training. This leads to efficient usage of resources and faster training. In order to have a meaningful representation of $I_q$ instances in the mini-batch we needed to ensure that a minimal number of that identity images are represented, i.e. if we take $I_q$ as cat image, then we need to have enough variations of cats and its augmentations in the same mini-batch to create good *argmax* hard positive pairs. Additionally, random sample of negative images (dis-similar images) are added to each mini-batch. Negative images are selected from within the crawled repository and app icons. Instead of picking the hardest positive pair, we use all positive pairs in that particular mini-batch while still selecting the hardest negatives adhering to the pair constraint in Eq. 2. However, in practice when hardest negatives are introduced early on in the training it leads to bad local minima thus leading to a collapsed model. Instead of introducing hardest negatives early on in the training, we adopted ideas from Curriculum learning which help in faster model convergence by presenting the model tougher concepts later on in the training process.

```
Algorithm 1 Online Pair Mining Strategy OPMS
Require: numEpochs for network training
 1: step = 1/numEpochs
 2: λ = 0
 3: for e do numEpochs
 4:     λ+ = step
 5:     for b do getImageMiniBatches()
 6:         positivePairs = generatePositivePairs(b)
 7:         negativePairs = generateRandomNegativePairs(b)
 8:         curriculumSortedPairs = OPMS(positivePairs, negativePairs, λ)
 9:         train model
10:     end for
11: end for
12: procedure OPMS(positivePairs, negativePairs, λ samplingWeight)
13:     margin m = 1
14:     calculate positivePairs and negativePair L2 euclidean distances
15:     sort pairs on L2 distance
16:     prune pairs argmax(D(x_q, x_p)^2) + m < argmin(D(x_q, x_n))^2  Eq. (2)
17:     negativePairs = weightedRandomSample(negativePairs, λ)
18:     pairs = CurriculumSortPairs(positivePairs, negativePairs)
19:     return pairs
20: end procedure
```

*Algorithm1: Online pair mining strategy pseudocode*

**Curriculum Learning**: by choosing which examples to present and in which order to present to the learning system, one can guide the training and remarkably increase the speed at which learning can occur. This idea is routinely exploited in animal and human training where it is called *shaping* [8]. In our scenario, instead of choosing the hardest negative pairs at the beginning of a particular epoch training, we introduce such pairs at the end. i.e. the entropy of the training examples increase as training epochs increase. Specifically, the pairs are sorted on L2 distance from the anchor image $I_q$ such that simpler pairs (away from the margin *m*) are presented first and tough pairs (near the margin *m*) are presented at the end. This approach results in more stable model training while at the same time helping us achieve faster convergence. In our case, we used mini-batches of size around 2000. See Results section 6 for more details. The online pair mining strategy pseudo-code is shown as Algorithm 1.

**Naïve approach**: We also explored naïve approach where instead of using pair constraint and curriculum learning we simply pick random positive and negative image pairs. This results in an inferior model as shown in Results section 6.

*B. Multi-Scale CNN*

We used a *CNN* architecture that employs different levels of invariance at different scales, inspired by [10, 49]. Our goal was to have a high-quality image embedding. *CNN's* have shown to achieve good performance for image classification [47]. They are able to do this by encoding strong invariance in its architecture which is learnt during training. This invariance is generally higher at top layers of a *CNN* but this invariance can be harmful for fine-grained image similarity tasks like ours as the final embedding might not encode the simpler aspects of an image like shape, colors etc. With the architecture shown in Fig. 5 we hope to extract a better image embedding for similarity tasks.

The *CNN1* in *fig*.5 has the same architecture as the convolutional deep neural network in *VGG16* [9]. The *CNN1* encodes strong invariance and captures the image semantics as this model has 16 convolutional layers and the top layer has the complex image feature representation. The other two parts of the network *CNN2* and *CNN3* takes down-sampled images and use shallower network architecture. These two parts have less invariance and capture the visual appearance (simpler aspects of the images like color, shapes etc.). Three different *CNN's* were employed instead of using one *CNN* and sharing lower layers as this way each *CNN* architecture can be evolved independent of the other two.

Finally, we normalize the embedding's from the three *CNN's* and combine them with a linear embedding layer. The final joint image embedding is represented as a 4096D vector. We use *L2* normalization to prevent overfitting. Results in section 6 show that this custom multi-scale *CNN* architecture outperforms single scale *CNN* for image similarity task. A brief note on combining embedding's across multiple sub-spaces. When having a deep network *CNN1* with high entropic capacity, having a high dimensional final layer (here *4096D*) allows information to be effectively encoded in this sub-space. For shallower network architectures like *CNN2* and *CNN3* we can get away with less dimensions (here *512D* and *1024D* resp.) as higher dimensional sub-spaces will mostly be sparse.

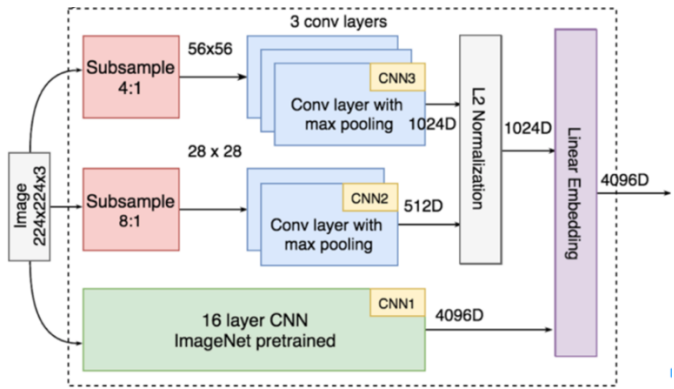

*Fig. 5 Multi-scale CNN architecture inspired from [49]. Each CNN from fig.2 is composed on the above multi-scale architecture. Three CNN's each learning features at different scales. Finally, a joint embedding of top as well as bottom layer features is used as embedding. Two such CNN's are trained together with shared weights in Siamese style.*

### IV. SIMNET MODEL TRAINING

We briefly present here some of our learnings while training a complex system like *SimNet* end-to-end. Detecting and preventing model overfitting is one of the main concerns especially when employing online pair mining strategy where we are not exposing the model to all possible pairs.

**Training:** using of pre-trained models via Transfer learning [12] helps in faster model convergence. In our case, *VGG16* was used which was pre-trained on ImageNet data. Fine-tuning a pre-trained *CNN* was done with a very slow learning rate. *SGD* optimizer was used [14] rather than an adaptive learning rate optimizer such as *RMSProp* [15]. This is to make sure we have control over the magnitude of the updates. Learning rate, decay and momentum was experimented with which helps the optimizer to continue making updates in the right direction when learning rate shrinks to small values. This helps prevent the model to be stuck in local minima. For a pre-trained network, the updates should stay very small so as not to wreck the previously learned weights. Learning rate (*lr*) is the step size during forward/backward propagation in neural network training. Setting the right *lr* is crucial for convergence. When training with a particular *lr,* training loss should decrease very quickly at the start of an epoch. If this is not happening, it's advisable to stop training and adjust *lr*.

Since there could be noise in training data, choosing the right number of training epochs is important for model convergence without overfitting. During cross validation, the validation loss between epochs give valuable information about the model training. With a good number of epochs and learning rate; as training goes to deeper epochs there should be a decreasing trend of validation loss with minor fluctuations. That is a clue to stop training. Experiment to select appropriate model hyper-parameters – number of layers, number of convolution filters, size of each layer, stride dimensions, padding, learning rate, type of optimizer. Also, look at debugging section which gives an intuition about how the model is training. Images data was preprocessed to unit mean and normalized to achieve faster model convergence and training.

**Overfitting:** Augmentation of images with random transformations (no image is seen twice) helps *SimNet* become more robust and prevent overfitting. Dropout [16] has shown to prevent overfitting. Dropout prevents a layer from seeing the exact same pattern twice, thus acting analogous to data augmentation. In a way, both dropout and image augmentations tend to disrupt random correlations occurring in your data. It has also been shown that dropout is essentially equivalent to *L1* norm [17] thus preventing overfitting. This was used when merging embedding's across different *CNN* sub-spaces in multi-scale *CNN*. Fine-tuning a pre-trained *CNN* can be tricky work. Depending on the volume of data, it can be appropriate to fine-tune all layers or only top few layers. Fine-tuning all layers with less data leads to an overfit model [26, 27]. In our case, only the top 2 convolution layers was fine-tuned. This aspect has to be experimented while training.

**Testing:** Test set was not polluted for hyper-parameter tuning. In our case, 5-fold cross validation was done to decide on hyper-parameters. Test set was used only once at the end to give us our generalized performance. This is what is reported in the Results section 6.

**Process:** This trained model compares any new incoming image against a repository of images. Matches are ranked on L2 distance and the top *n* are used. On a GPU host, it takes a on avg. 3 seconds to evaluate each new image on 10K repository of category test images. If category repository gets really large this could be further optimized depending by using more powerful GPU's or narrow down the scope of search using locality sensitive hashing methodologies [38, 55].

**Debugging**: plays a crucial role in understanding how and what the model is learning. As a debugging methodology, it's beneficial to see image visualizations of the activated neurons across layers to get an intuition of what the model has learnt. This helped better understand the mistakes the model is making. In certain cases, it's a good practice to perform gradient checking to see if backpropagation is working fine i.e. are the gradients flowing across the layers, what is the magnitude of gradient updates as training epochs are increasing. This is to cross check model learning. Plotting layer weights, train and test loss per epoch, precision-recall curves, roc-auc curves are all invaluable.

## V. RELATED WORK

Image similarity has long history in information retrieval and prior to that in storage systems (databases). Image similarity has broadly been explored using 1) Image Content – using image content to find similar, 2) Text based – text surrounding the image to understand image, 3) Semantic, 4) Sketch based – use input as sketches using which relevant images are retrieved and 5) Annotation based approaches [43]. The primary computation approach used in all the above approaches is the same – collect a database of images and store them. Have a function which can compute similarity between any two images. At runtime, given a new image, similar images are retrieved from storage. Prior art in image similarity exists in efficient ways to crawl and gather reference images, appropriate ways to computing similarity, ways to compute similarity efficiently and fast.

Traditionally visual features were heuristically designed as local and global features [50-54] using color, shape. texture in images. SIFT, SURF, ORB [45] also were popular ways to compute image similarity.

Image similarity using Convolutional networks were explored by *LeCun* et al in Handwritten recognition where the task was to retrieve / recognize a digital image [46] using Siamese networks. The current (circa 2016-17) Deep CNN's performance out-shines previous approaches due to requiring no hand-tuned features. The CNN's automatically learn a representation of the image based on the objective function, provided data and the network architecture. This gives better model generalization performance [47] in practice. *Melekhov* et al [48] have explored the use of Siamese networks for image matching. *Babenko et al*. [28] showed how domain adaptation using transfer learning can improve performance in practice. There have also been early approaches (circa 2010-2011) inter-mixing non-deep learning approaches like Fisher vectors with CNN features [30] leading to improved performance compared to non-deep learning approaches. But as deeper and more expressive CNN models came into existence [62-64] to the best of our knowledge feature inter-mixing based explorations (like [30]) does not happen.

Deep ranking [49] proposes a similarity metric-learning directly from images using a triplet network and using Deep Siamese CNN's [58]. There have also been works related to image similarity using image *patches* [32-37, 55-57], we do not explore more in this direction as image similarity is more challenging than patch matching due to image similarity dealing with bigger image sizes (thereby having to deal with viewpoint, appearance, lighting, potential distortions).

Metric learning and pair mining is explored in a related work using triplet neural networks in FaceNet [21]. Hard-pair mining is also explored in [59-61]. But our approach with OPMS (algorithm 1) and multi-scale CNN's (fig.5) promises orders of magnitude faster convergence on our data.

Curriculum learning (CL) [4] is a methodology to train ML models by introducing easier training examples first and gradually increasing the difficulty level of the examples. Motivation for this type of learning comes from the observation that humans and animals often seem to learn better and faster when trained with a curriculum like strategy.

In our work we focus on Content based similarity using images only. While we use deep learning models to better represent images in embedding space, we draw attention to effective and faster ways to train such models using Curriculum learning. We show experimentally that Curriculum learning strategies significantly decrease model convergence time with little sacrifice in performance.

## VI. RESULTS AND IMPACT

Models are evaluated on a test dataset. The test set has the same 10 categories and contains 10K images, 1K per category. As can be observed, the class distribution in this test data was similar to train data. This ensures we get to measure approx. generalized performance of the model. The main use of our projection $I_q \rightarrow x_q$ (image $q$ to embedding of $q$) is to look up visually similar images. Note that the query image $I_q$ in real-life could be significantly deformed, rotated, occluded, scaled.

**Embedding space visualization**: To visualize this, see Fig. 6. The final 4096D embedding space (trained with architecture F see table 1) is projected to 2D using t-SNE [13]. Different variations of category image (some with drastic modifications to color and deformations are grouped nearby. One can see that in general *SimNet* does a good job of projecting similar images close by in the embedding space. This is possible as a result of multi-scale CNN learning a higher quality image embedding.

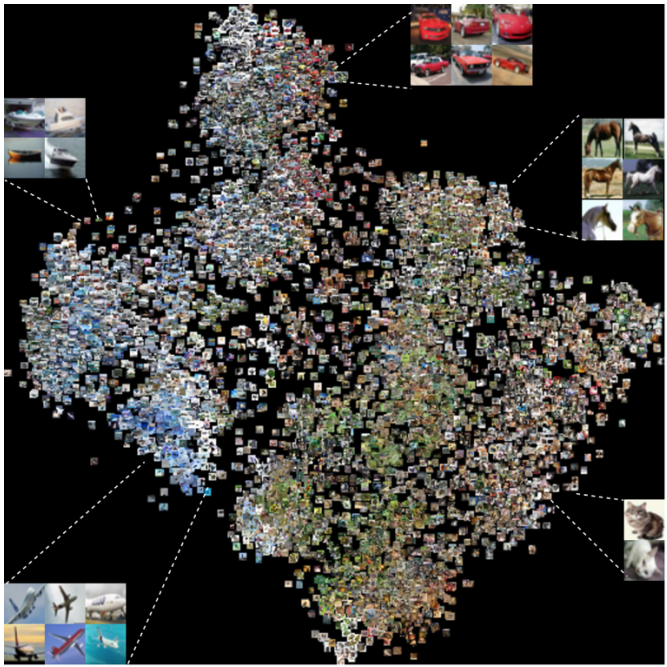

*Fig. 6 t-SNE 2D embedding space visualization of images. Embedding obtained from model F (table 1).*

**Embedding dimensions:** We also studied the effect of embedding dimension on the final performance. We tried 512, 1024, 2048, 4096*D* embedding's to represent each image projection $I_i \rightarrow x_i$. For shallower *CNN's* architecture, 512D and 1024 D embedding space was sufficient as high dimensional embedding's were not improving performance. For deep *CNN*, 4096D embedding was used. Also, the joint embedding from all three *CNN's* has 4096D embedding.

**Evaluating metrics**: All trained models are evaluated in terms of accuracy@1. We used contrastive loss function as described in section 3. Baseline model was an ImageNet pre-trained *VGG16 CNN*. Table 1 shows results from training different models.

| Model | Accuracy |
|---|---|
| A. Random guess in data | 10% |
| B. ImageNet pre-train classifier VGG16 - baseline | 90.2% |
| C. Fine-tune; pre-train classifier VGG16 | 91.2% |
| D. ImageNet pre-train similarity VGG16 - baseline | 90.28% |
| E. Siamese network; fine-tune pre-train VGG16 | 91.9% |
| **F. Multi-scale Siamese fine-tune pre-train VGG16 with OPMS 5.5 million pairs** | **92.6%** |

*Table 1. Different model performance comparisons. A is a random guess in the catalog. B and D are baseline classifier and similarity models. C, E are fine-tuned models on our data. F is SimNet – fine-tuned multi-scale CNN Siamese network trained with OPMS.*

**Online pair mining:** here we present the impact of online pair mining strategy (*OPMS*) on final model performance and training time. The main reason for this training strategy was based on the intuition that not all image pairs lead to equal model learning. Additionally, structured learning inspired by curriculum learning, would help the model learn better, faster and with far fewer examples. Table 2 show results which validate this hypothesis.

| Pair formation method | # of pairs (million) | Accuracy | Train (days) |
|---|---|---|---|
| A. OPMS with λ=0.2 | 1.1 | 81.0% | 1 |
| **B. OPMS with λ=0.5** | **5.5** | **92.6%** | **3** |
| C. OPMS with λ=0.7 | 10.1 | 90.2% | 6 |
| D. OPMS with λ=0.9 | 15.7 | 90.4% | 7 |
| E. OPMS with λ=1.0 | 21.3 | 92.8% | 12 |
| F. naïve approach - random | 5.5 | 78.1% | 3 |
| G. All pairs | 500 | - | ≈378 |

*Table 2. Impact of online pair mining strategy*

From table 2, we can see *OPMS* with different λ values. This controls the selection of hard positive and negative pairs for training as per Algorithm 1. With λ=0.2 we see that as not enough (tough) pairs were part of training, the Accuracy is only 81%. On the other end of the spectrum when λ=0.7, 0.9 we see the Accuracy increase but not by much. This goes on to show that probably learning has stagnated. Intuitive, when λ=1.0 and all pairs are selected (include hardest pairs), the precision is best at 92.8%. Considering all pairs certainly gives the best accuracy on CIFAR 10 data, but the same might not be true on noisy datasets. The intuition being the noisy and mislabeled data are the hardest pairs being present near the decision boundary. Considering all such pairs could decrease model performance.

The best model on CIFAR 10 is the one trained on all pairs at λ=1.0 on 21.3 million pairs it took 12 days to train. On the contrary, the model trained at λ=0.5 on 5.5 million pairs took only 3 days to train while giving little decrease in model performance (drop of 0.2% accuracy). i.e. due to online pair mining strategy we had model convergence 4X quicker with only 0.2% drop in accuracy. Also, note the naïve approach in F, where selecting 5.5 million random pairs (same pairs as winning model λ=0.5), the model performance drops significantly to 78.1% indicating the benefit of *OPMS* in model convergence and performance.

**Runtime:** Current *CNN's* have long training times but very short prediction time. On a AWS p2.xlarge instance with *GK210 GPU's*, we can compute $I_i \rightarrow x_i$ for any image in about 55ms (TP90) with most of the time spent loading the image. In Siamese network with the weight matrix shared between the *CNN's* the model has less number of parameters to learn. As a future iteration, we are planning to replace *VGG16* (which has about 140 million parameters) with *ResNet* [18] or *GoogLeNet* [19] which have almost 20X less parameters. With this change, we expect the train time to reduce and model performance to further improve.

**Final model hyper-parameters:** For the best model F, we present the hyper-parameters which we got via 10-fold cross-validation on training data. We used *SGD* optimizer with learning rate of 1e-5 with decay of 1e-6 and nesterov [20] momentum of 0.9. We augmented images to prevent overfitting. Image augmentation involves - random crops, rotations, zoom, adding gaussian noise, affine transforms, changing channel-wise contrast and brightness; all of which done in random order. We fine-tuned the top two conv layers of pre-trained *VGG16 CNN*. We applied 0.5 dropout in fully-connected layers. ReLU activation was used in conv layers to learn non-linearity in data. (1, 1) 2D Zero padding was done to the image borders. Training was done for about 100 epochs on about 5.5 million pairs as mentioned in table 2. When training Siamese network, a few versions of distance metrics was tried – Euclidean, cosine, L1 (city-block) distance. We found Euclidean was doing as well as or better than the other metrics. Finally, only Euclidean distance metric was used.

## VII. SUMMARY

We have presented *SimNet* to perform image similarity given a reference image. We achieved this by training a multi-scale Siamese *CNN* is better at finding fine-grained image similarities than traditional CNN's. We also presented a novel online pair mining strategy inspired by Curriculum learning to help in faster model training and convergence almost 4x faster with minimal impact on model performance.